# Reinforcement Learning Based Handwritten Digit Recognition with Two-State Q-Learning

**Abdul Mueed Hafiz*[1], Ghulam Mohiuddin Bhat[2]**

[1, 2] Department of Electronics and Communication Engineering
Institute of Technology, University of Kashmir
Srinagar, J&K, India, 190006.

```
Email(Corresponding Author)*: mueedhafiz@uok.edu.in
      Email (Co-author): drgmbhat@uok.edu.in
```

ORC-ID[1]: 0000-0002-2266-3708
ORC-ID[2]: 0000-0001-9106-4699

**Abstract**

We present a simple yet efficient Hybrid Classifier based on Deep Learning and Reinforcement Learning. Q-Learning is used with two Q-states and four actions. Conventional techniques use feature maps extracted from Convolutional Neural Networks (CNNs) and include them in the Q-states along with past history. This leads to difficulties with these approaches as the number of states is very large number due to high dimensions of the feature maps. Since our method uses only two Q-states it is simple and has much lesser number of parameters to optimize and also thus has a straightforward reward function. Also, the approach uses unexplored actions for image processing vis-a-vis other contemporary techniques. Three datasets have been used for benchmarking of the approach. These are the MNIST Digit Image Dataset, the USPS Digit Image Dataset and the MATLAB Digit Image Dataset. The performance of the proposed hybrid classifier has been compared with other contemporary techniques like a well-established Reinforcement Learning Technique, AlexNet, CNN-Nearest Neighbor Classifier and CNN-Support Vector Machine Classifier. Our approach outperforms these contemporary hybrid classifiers on all the three datasets used.

## 1. Introduction

Handwritten digit recognition [1-5] is an interesting research problem. Many techniques have been proposed for this problem, such as deep learning [6-10] based classification techniques [11-16], artificial neural networks [17,18], and support vector machines [19,20]. Although many of these techniques have achieved satisfactory classification accuracy, challenges remain due to several issues including non-standard writing habits in circle and hook patterns, e.g. in 4, 5, 7 and 9. Recently Reinforcement Learning has been applied to this area generating promising results [21-23]. Reinforcement Learning [24,25] is garnering much attention in research [26-30]. In the field on computer vision, impressive initial work [31-36] has been done. Taking a cue from the work done in real-time tracking of Maximum Power Point Tracking (MPPT) done in Photovoltaic Arrays [29], we propose a Hybrid Classifier for handwritten digit recognition technique giving high accuracy which is based on Deep Learning and Reinforcement Learning. By using features generated from last fully-connected layer of the trained Convolutional Neural Network (CNN), and also by using Reinforcement Learning, the technique proposes an action (in our case rotation by a certain angle) on the image. After applying the action to the original image, classification is done.

Most of the Reinforcement Learning based techniques [36,35,34,33,32,31,37,38] used for object detection in images use actions like zooming and translation as per human visual comprehension/detection. However, they miss an important action of rotation which is also used by humans in visual comprehension/detection. We use rotation in our proposed approach which is a first in this regard as per our knowledge. We use Q-Learning [39,40] based Reinforcement Learning. The Q-states used in the conventional Reinforcement Learning based Object Detection/Digit Recognition techniques use features with large dimensions along with history of states. This leads to very large state space leading to difficulty in optimization. Our technique just uses two states and four actions. Hence the Q-table has two rows and four columns. In spite of the simplicity of our technique, it achieves better results as compared to other contemporary techniques like Deep-Reinforcement Networks [23], Hybrid Classifiers based on Convolutional Neural Network (CNN) viz. CNN-Nearest Neighbor [41-43] and CNN-Support Vector Machine [44].

## 2. Proposed Approach

We propose a hybrid approach of deep learning and reinforcement learning. First a naive Convolutional Neural Network (CNN) is trained on the dataset. The architecture of the CNN used is showed in Figure 1. Next, the feature map of the first fully connected layer of the CNN with input: 576 and output: 3136, is extracted for each training sample by feeding the sample to the trained CNN. These extractions are referred to as the fc-Layer emissions, and are collectively referred to as $F_{Train}$. The feature vector for one image is an array of size 1-*by*-3136 . For classification of a test sample it is first fed to the trained CNN. Next, the fc-layer emission of the sample ($F_{Sample}$) is extracted. Then Nearest Neighbor based Exhaustive Searching is done for $F_{Sample}$ over entire $F_{Train}$ giving Nearest Neighbor Distance $D$. Based on a selection criterion, if the sample is tagged as 'hard to classify', it is classified using Reinforcement Learning, otherwise its class is assigned as that of the Nearest Neighbor. This paper is not about selection criteria and we do not use any such criteria except that all samples wrongly classified by the Exhaustive

Searcher are tagged as 'hard to classify.' The Reinforcement Learning based classification is as follows. A randomly selected permutation is applied to the test image. Then the new fc-layer emission of the permuted image ($F'_{Sample}$) is extracted from the trained CNN.

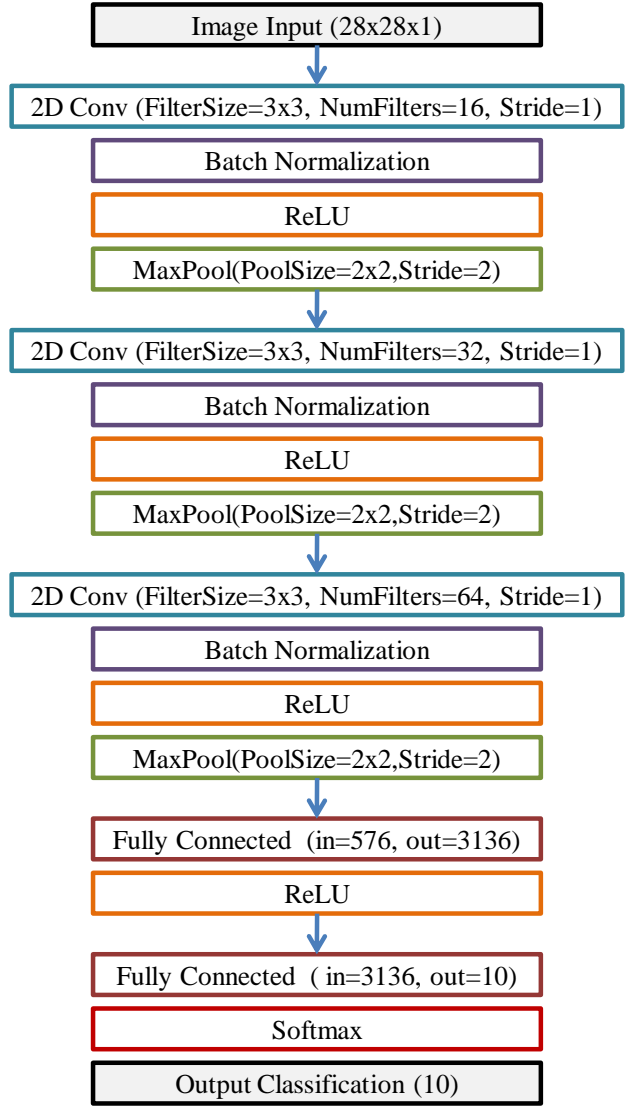

Figure 1. Architecture of the CNN used

The permutation used here is image rotation with a randomly selected angle from a set (**A**) of four different angles = $[-15°, -10°, +10°, +15°]$. Then, Nearest Neighbor based Exhaustive Searching is done for $F'_{Sample}$ over entire $F_{Train}$ giving Nearest Neighbor Distance $D_1$. For Reinforcement Learning based decision making, Random Policy Based Q-Learning is used. Two states *(n = 2)* and four actions *(a = 4)* are used. The current state is decided after observing whether $D < D_1$ or $D > D_1$. The four actions are based on applying each of the four rotation angles. The Q-Table which has 2 rows *(n = 2)* and four columns *(a = 4)* is initialized by making all entries as zeros. The reward $r$ which is based on the difference $\Delta D = D_1 - D$, is given as follows:

$$r = \begin{cases} -1, & if\ \Delta D > \delta \\ 0, & if\ -\delta < \Delta D < \delta \\ +1, & if\ \Delta D < -\delta \end{cases}, \text{ where } \delta = 0.05 \qquad (1)$$

The number of iterations for updating the Q-Table is based on $N = a \times m$, where $a=4$ and $m=30$. After each iteration the Q-value for the 'state-action pair entry' with state $s$ and action $a$, viz. $Q(s, a)$ in the Q-Table, is updated according to Q-Learning Update Rule as:

$$Q(s,a) = Q(s,a) + \alpha[\ r + \gamma \max_{\forall b \in A} Q(s',b) - Q(s,a)\ ] \qquad (2)$$

Where $s'$ is the new state, $\alpha = 0.3$ is the learning rate, and $\gamma = 0.6$ is the discount rate. The flowchart for the proposed Reinforcement Learning (RL) Algorithm is shown in Figure 2. After the completion of $N$ iterations of Q-Learning, the optimal action is chosen for that action which has highest Q-value in the Q-Table. The optimal action is then applied to the original test image. The image is then fed to the trained CNN and the fc-Layer emission is obtained. This feature vector is fed to the Nearest Neighbor Exhaustive searcher which searches for the nearest neighbor over $F_{Train}$. Finally the test image is assigned the class of the nearest neighbor. Figure 3 shows the proposed approach.

## 3. Experimentation

Experimentation was done on an Intel® CORE i3® processor machine with 6GB RAM running Windows 7. For comparison of the performance of the proposed approach, various other contemporary approaches were also implemented. These were AlexNet [45], a Deep Q-Network (DQN) based approach [23], *CNN-Nearest Neighbor* Hybrid Approach [41] and *CNN-Support Vector Machine* Hybrid Approach [44]. The approach proposed in [23], uses a Deep Q-Network based reinforcement learning approach for classification of digits and is quite relevant to the work at hand. Also, the relevant hybrid approaches consisted of extracting feature-maps of first-from-start fully connected layer in the trained CNN and classifying the features using Nearest Neighbor (NN) Classifier, or Support Vector Machine (SVM) Classifier respectively. Benchmarking was done on three digit image databases viz. *MNIST* [1], *USPS* [46] and *MATLAB Digit Image Database* [47]. Figure 3 shows the architecture of the CNN used. The dimension of the features extracted from the last fully-connected layer of the trained CNN was 1×3136. The CNN was trained using *Learning Rate* = 0.01, *Maximum Epochs* = 1, *Mini-batch Size* = 40, *Validation Frequency* = 50. Most of these parameters are MATLAB defaults for training CNNs. The data images used were uniformly distributed and randomly selected as follows: *Training Classes* = 10, *Training Images* = 6000, *Validation Images* = 2000, and *Testing Images* = 2000. Table 1 shows the performance of various approaches on each of the datasets respectively.

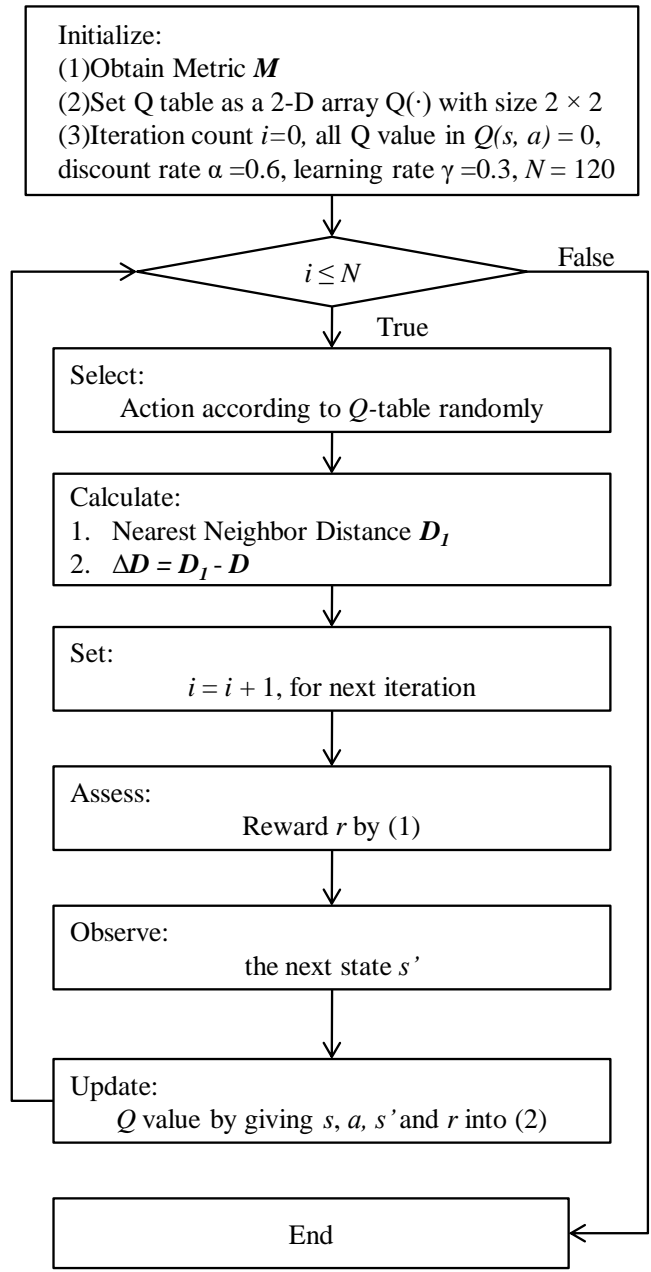

Figure 2. The flowchart of the proposed Reinforcement Learning Algorithm

Table 1. Performance Benchmarking of Various Approaches

| Approach | Classification Accuracy On MNIST | Classification Accuracy On USPS | Classification Accuracy On MATLAB Digit Dataset |
|---|---|---|---|
| **Alexnet [45]** | 96.6% | 97.3% | 96.5% |
| **CNN-Nearest Neighbor Hybrid [41]** | 97.6% | 99.4% | 100.0% |
| **CNN-SVM Hybrid [44]** | 98.0% | 99.2% | 99.9% |
| **DQN based approach [23]** | 98.6% | 99.5% | 100.0% |
| **Proposed Approach** | **99.0%** | **99.7%** | **100.0%** |

As is observed from Table 1, our approach generally outperforms the other approaches on all of the three datasets. We do not use dimensional reduction or network training, and our approach is instance-based. Hence, in this work, the processing time is larger.

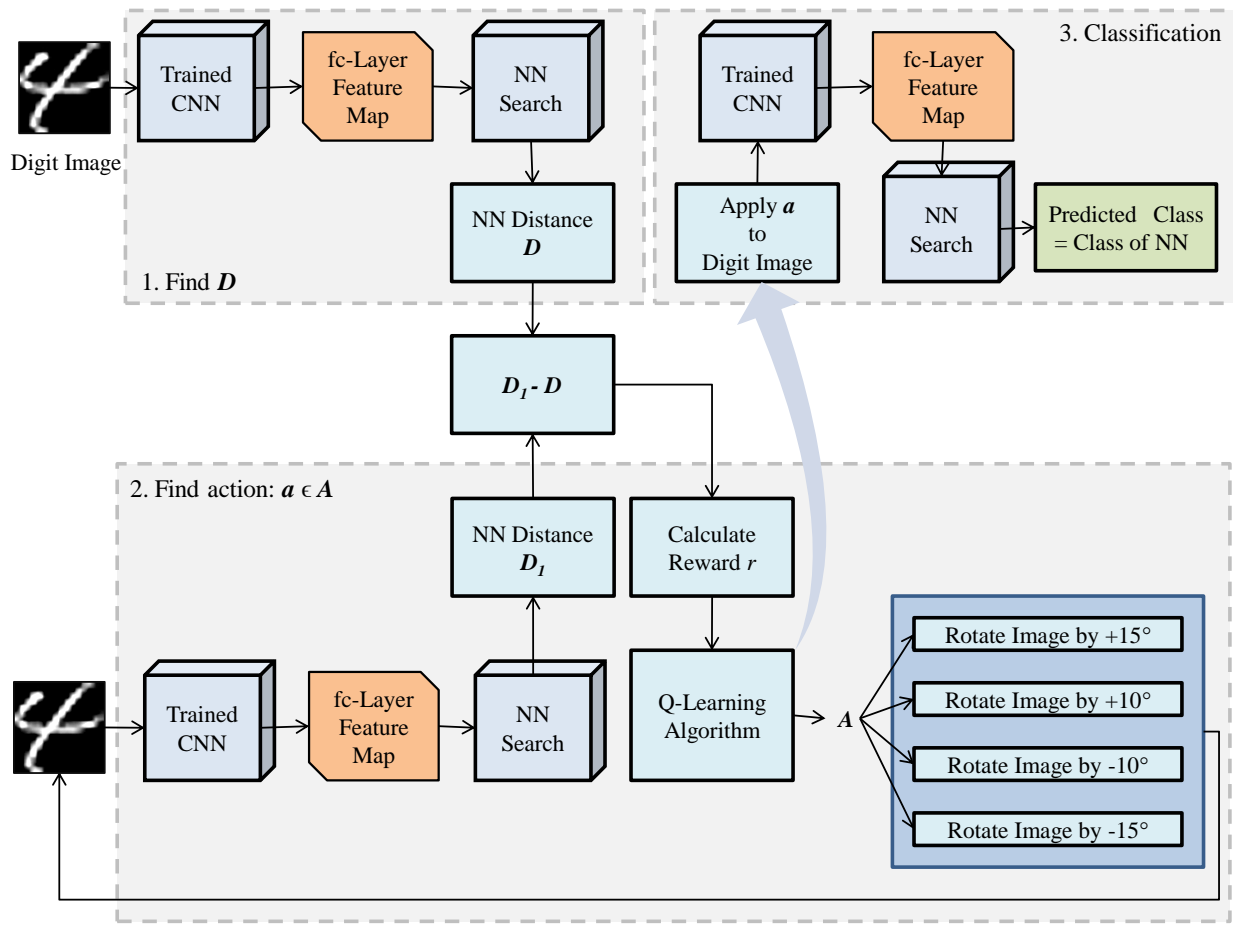

Figure3. Proposed Approach ('NN' means' Nearest Neighbor')

## 4. Conclusion and Future Work

In this paper we investigate a simple and efficient hybrid classifier combining deep learning with Q-Learning based Reinforcement Learning. We say that the proposed approach is simpler than other techniques found in literature because unlike the latter using large number of states, the former uses only two states. Hence, it is easy to optimize and has a straightforward reward function. Contemporary techniques use vision tasks like zooming and translation. However, to the best of our knowledge the proposed approach is the first one to use rotation which is similar to tilting of vision. Three datasets were used in the course of the experimentation. These were MNIST, USPS and MATLAB Digit Image Dataset. The performance of the proposed hybrid classifier was compared with that of other conventional classifiers (e.g. AlexNet) and also other contemporary hybrid classifiers. These were Deep-Reinforcement Learning Network, CNN-Nearest Neighbor and CNN-SVM. Our approach generally outperformed these classifiers on all the three datasets. Future work would involve making the approach faster e.g. by using dimensional reduction or by using a smaller size of the Fully Convolutional Layer / feature-map. Future work would also include using the approach in other areas of computer vision like object detection i.e. on a larger/more comprehensive scale.

## Conflict of Interest Statement

The authors declare no conflict of interest